\documentclass[10pt,twocolumn,letterpaper]{article}
\usepackage[pagenumbers]{cvpr} 

\usepackage{graphicx}
\usepackage{amsmath}
\usepackage{booktabs}
\usepackage{microtype}
\usepackage{tabularx}
\usepackage[accsupp]{axessibility}

\usepackage[pagebackref,breaklinks,colorlinks]{hyperref}

\newcommand{\R}{\mathbb{R}}

\newcommand{\comment}[1]{}

\newcommand{\vd}{\mathbf{d}}

\newcommand{\vk}{\mathbf{k}}

\newcommand{\vp}{\mathbf{p}}

\newcommand{\vv}{\mathbf{v}}

\newcommand{\mD}{\mathbf{D}}

\newcommand{\mH}{\mathbf{H}}
\newcommand{\mI}{\mathbf{I}}

\newcommand{\mS}{\mathbf{S}}

\newcommand{\mV}{\mathbf{V}}

\newcommand{\cL}{\mathcal L}

\newcommand{\cP}{\mathcal P}

\newcommand{\cT}{\mathcal T}

\def\cvprPaperID{7055} 
\def\confName{CVPR}
\def\confYear{2023}

\begin{document}
\title{Few-shot Geometry-Aware Keypoint Localization}

\author{Xingzhe He$^1$\thanks{Work was done while interning at Flawless AI}
\and
Gaurav Bharaj$^2$
\and
David Ferman$^2$
\and
Helge Rhodin$^1$
\and
Pablo Garrido$^2$
\and
\begin{tabular}{ c c }
\small{$^1$ University of British Columbia} &\small{$^2$ Flawless AI}
\end{tabular}
}
\maketitle

\begin{abstract}
Supervised keypoint localization methods rely on large manually labeled image datasets, where objects can deform, articulate, or occlude. However, creating such large keypoint labels is time-consuming and costly, and is often error-prone due to inconsistent labeling. Thus, we desire an approach that can learn keypoint localization with fewer yet consistently annotated images. To this end, we present a novel formulation that learns to localize semantically consistent keypoint definitions, even for occluded regions, for varying object categories. We use a few user-labeled 2D images as input examples, which are extended via self-supervision using a larger unlabeled dataset. Unlike unsupervised methods, the few-shot images act as semantic shape constraints for object localization. Furthermore, we introduce 3D geometry-aware constraints to uplift keypoints, achieving more accurate 2D localization. Our general-purpose formulation paves the way for semantically conditioned generative modeling and attains competitive or state-of-the-art accuracy on several datasets, including human faces, eyes, animals, cars, and never-before-seen mouth interior (teeth) localization tasks, not attempted by the previous few-shot methods. Project page: \href{https://xingzhehe.github.io/FewShot3DKP/}{https://xingzhehe.github.io/FewShot3DKP/}
\end{abstract}

\section{Introduction}
\label{sec:intro}

\begin{figure}
    \centering
    \includegraphics[width=\linewidth]{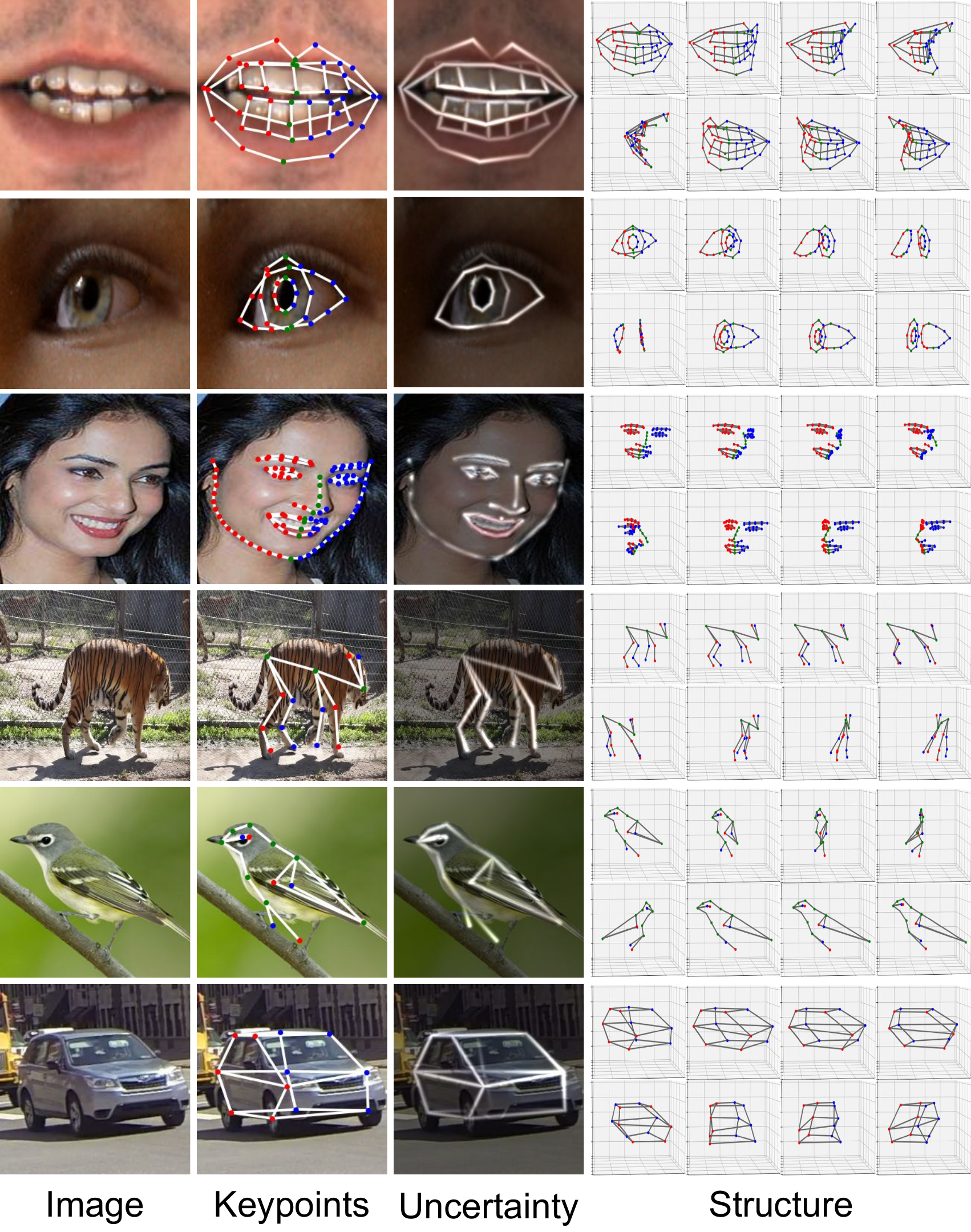}
    \caption{All results are obtained by 10-shot learning except Tigers where 20 examples are used. The left/right/middle keypoints are marked in red/blue/green. Using only a few shots, the model learns semantically consistent and human interpretable keypoints. Uncertainty modeling helps us identify occlusions and ambiguous boundaries, as shown in the mouth, eye, and car example. 
    }
    \label{fig:teaser}
\end{figure}

Keypoint localization is a long-standing problem in computer vision with applications in classification \cite{csurka2004_vis-categor, choutas2018_potion}, image generation \cite{ma2018_disentangled-person, siarohin2018_deformable-gans}, character animation \cite{siarohin2019_monkeynet, siarohin2019_fomm}, 3D modeling \cite{feng2018_3dface-recon, pavlakos2019_3drecon}, and anti-spoofing \cite{deb2020_anti-spoofing}, among others. 
Traditional supervised keypoint localization approaches require a large dataset of annotated images with balanced data distributions to train robust models that generalize to unseen observations \cite{he2017_maskrcnn, wei2016_cpm, wood2022_dense-landmarks}.
However, annotating keypoints in images and videos is expensive, and usually requires several annotators with domain expertise \cite{wang2016_cephalometric-xray, wang2019_pelvic-xray, lu2020_hand-xray}. Manual annotations can be inaccurate due to low resolution imagery \cite{chandran2020_attention-driven} and temporal variations in illumination and appearance \cite{jin2020_whole-body, wandt2021_canon-pose}, or even subjective, especially in presence of external occlusions \cite{kumar2020_luvli, wu2018_wflw} and image blur effects \cite{sun2019-fab, zhang2020_blur-countering}. 
Besides, modeling self-occluded object parts is proven to be an ambiguous task since 3D consistent keypoint annotations are needed \cite{zhang2018_joint-voxel}. 
As a consequence, supervised approaches are prone to learning suboptimal models from noisy training data.
\looseness=-1

Unsupervised keypoint detection methods can predict consistent keypoint structures \cite{he2022_autolink, zhang2018_unsupervised, he2022_ganseg, he2021_latentkeypointgan, jakab2018_unsupervised, lorenz2019_unsupervised}, but they lack human interpretability or may be insufficient, e.g., for editing tasks
requiring detailed manipulation of object parts.
Jakab et al.~\cite{jakab2020self} pioneered adding interpretability with a cycle loss between unsupervised images and unpaired pose examples. However, their focus is different and the proposed cycleGAN struggles in a few-shot setting as it does not exploit paired examples. Unsupervised methods can be extended to few-shot setups either by learning a mapping that regresses detected keypoints to human-labeled annotations \cite{thewlis2017_unsupervised}, which requires hundreds or thousands of examples, or by attaching few-shot annotated examples to the unsupervised training batch as weak supervision \cite{moskvyak2021_semi-supervised}. However, we show that neither approach produces competitive predictions when added to state-of-the-art unsupervised keypoint localization methods. 
Our contributions focus on making the latter approach work, building upon the unsupervised reconstruction in \cite{he2022_autolink} and the skeleton formulation in \cite{jakab2020self}.

Recent advances in semi-supervised keypoint localization has shown significant progress in the field. Still, most existing methods are 
specialized
for a single object category such as faces \cite{qian2019_aggregation, browatzki2020_3fabrec, wei2021_few-shot} and X-rays \cite{chen2022_semi-supervised, yin2022_one-shot, zhou2021_scalable, yao2021_one-shot}, or require hundreds or thousands of annotated examples to achieve competitive performance \cite{moskvyak2021_semi-supervised, wang2022_pseudo-labeled, pavllo2019_3d_human}. Our approach, however, only needs a few dozens of examples.
In an orthogonal direction, Honari et al. \cite{honari2018_semi-supervised} assist keypoint localization via equivariance transforms and classification labels, though the latter are not always available. Generative image labeling has also shown great promise \cite{zhang2021_datasetgan,wu2022_synthetic, sushko2022_one}. 
However, annotating StyleGAN-generated images is prone to artifacts and noise. 
Besides, generative approaches are limited to the underlying data distribution biases~\cite{tewari2020_stylerig, or-el2022_stylesdf}, thus decreasing overall keypoint localization performance.

Current limitations create the need for an approach that can leverage a smaller yet semantically consistent corpus of human labeled annotations while generalizing to a much larger unlabelled image set.
This paper presents a novel formulation that learns to localize semantically consistent keypoint definitions, even for occluded regions, for various object categories with complex geometry using only a few user-labeled images.
We use as input a few example-based user-labeled 2D images with predefined keypoint definitions and their linkages to learn to localize keypoints. Unlike unsupervised methods, the user-selected few-shot images act as semantic shape constraints for human-interpretable keypoint localization. To enable generalization to the target data distribution, we extend our approach via self-supervision using a larger unlabeled dataset. In addition, we introduce 3D geometry-aware constraints to model depth and uplift 2D keypoints in 3D with viewpoint consistency, thus achieving more accurate 2D localization.


Experimental results demonstrate that our proposed approach competes with or outperforms state-of-the-art methods in few-shot keypoint localization for human faces, eyes, animals, and cars using a only few user-defined semantic examples. We also show the capabilities of our keypoint localization approach on a novel data distribution, specifically the mouth interior, which has not been attempted with previous few-shot localization approaches.
Thus, our novel general-purposed formulation paves the way for semantically conditional generative modeling with a few user-labeled examples. We hope it will enable a broader set of downstream applications, including fast dataset labeling, and in-the-wild modeling and tracking of complex objects, among others.

\noindent Our key contributions are summarized as follows:





1. A novel formulation for few-shot 3D geometry-aware keypoint localization that works on diverse data distributions. 

2. We introduce keypoint uncertainty and local 3D aware geometry constraints for better keypoint localization. 

3. We adapt techniques of transformation equivariance and image reconstruction from unsupervised methods.


4. Our approach enables flexible modeling of complex deformable objects and geometric parts, such as mouth interior, faces, eyes, cars, and animals via a few user examples with consistent semantic definitions.

\section{Related Work}
\label{sec:relatedwork}

\textbf{Supervised keypoint localization}
Supervised methods learn keypoint localization by leveraging a large corpus of human-labeled images for standard object categories \cite{yi2014_coco} or domain-specific classes, such as faces \cite{bulat2017far, koestinger2011_aflw, wu2018_wflw, zhang2018_joint-voxel}, eyes \cite{kothari2019_giw}, teeth \cite{wang2016_cephalometric-xray}, human bodies \cite{andriluka2014_2dpose, ionescu2014_human3.6m, mehta2017_mon-3d-human, jin2020_whole-body}, animals \cite{li2020_atwr}, and vehicles \cite{xiang2014_pascal3d+, reddy2018_car-fusion}, among others. Annotating images and videos is not only expensive but also prone to labeling errors \cite{mathis2020_primer-mocap, hoiem2012_diagnosing-error, ronchi2017_benchmarking-error} due to image downsampling \cite{chandran2020_attention-driven}, object occlusions \cite{kumar2020_luvli, wu2018_wflw}, image blur \cite{sun2019-fab, zhang2020_blur-countering}, and harsh appearance and lighting variations especially in video datasets \cite{wandt2021_canon-pose, jin2020_whole-body}. As such, supervised models render inaccurate for downstream tasks, often requiring statistical uncertainty modeling \cite{ferman2022_mdmd, li2021_human-pose} or robust regressors \cite{dollar2010_cpr} to achieve state-of-the-art performance.
A recent line of work resorts to problem-specific high-quality synthetic datasets, mainly for human bodies \cite{patel2021_agora}, faces \cite{wood2021_fakeit}, eyes \cite{wood2015_rendering-eyes, nair2020_riteyes} and teeth \cite{wood2022_dense-landmarks} to produce perfect semantic annotations, even for partially occluded object parts. Despite these efforts, creating a synthetic dataset with real-world data distribution remains a challenging and very laborious task, and methods trained on them still achieve near-competitive performance \cite{wood2021_fakeit, wood2022_dense-landmarks, nair2020_riteyes}. On the other hand, our approach can learn a general model that preserves semantic definitions from limited annotations and user constraints while still generalizing well to an unseen distribution via self-supervision.

\textbf{Semi-supervised keypoint localization}
We draw a line between semi-supervised keypoint localization, where hundreds or thousands of image annotations are needed, and few-shot keypoint localization, where the number is restricted to dozens. 
Qian et al. \cite{qian2019_aggregation} transfer style of labeled images to augment the face appearance distribution of the training set.
Dong and Yang \cite{dong2019_teacher-supervises} propose a teacher network to select pseudo labels generated by student networks. The methods above handle limited head pose variations. 
Wang et al. \cite{wang2022_pseudo-labeled} extend the pseudo label idea to general image distribution by exploiting reinforcement learning, which can be unstable and computationally expensive. 
Extra information, such as classification labels \cite{honari2018_semi-supervised, ukita2018_weakly-supervised}, multiview constraints \cite{feng2021_active}, and video frames \cite{pavllo2019_3d_human} can help in semi-supervised learning, but it is not always available for all datasets.
Mathis et al. \cite{mathis2018_deepcutlab} fine-tune a pose estimation network \cite{insafutdinov2016_deepercut} on hundreds of labeled images for constrained animal pose detection tasks under simple laboratory conditions.
Moskvyak et al. \cite{moskvyak2021_semi-supervised} directly learn from unlabeled images, akin to our method. They impose equivariance constraints between images and keypoints such that features extracted at keypoints remain invariant to linear transformations. 
These semi-supervised methods still require a labeled dataset that is one or two orders of magnitude higher than that of our 
approach.

\textbf{Few-shot keypoint localization} 
Most existing few-shot keypoint localization methods focus on specific domains, mostly faces \cite{browatzki2020_3fabrec, wei2021_few-shot} or medical X-ray images \cite{chen2022_semi-supervised, yin2022_one-shot, zhou2021_scalable, yao2021_one-shot}. Browatzki et al. \cite{browatzki2020_3fabrec} pre-train an auto-encoder on millions of faces and modify it to generate keypoints. Similarly, Thewlis et al. \cite{thewlis2019unsupervised} pre-train on tens of thousands of faces and toy roboarms. Wei et al. \cite{wei2021_few-shot} fine-tune a pre-trained face landmark detector for custom keypoint locations. In the medical domain, X-ray images often share common appearance and viewpoints. Thus, state-of-the-art methods constrain 2D keypoint deviations \cite{chen2022_semi-supervised} and the features extracted at keypoint locations \cite{yao2021_one-shot}, or resort to unsupervised registration \cite{yin2022_one-shot} to adapt the keypoints from the few-shot examples to the unlabeled images. Although these approaches show effectiveness for X-ray images, they may not be applicable to images with larger viewpoint and appearance variations. Instead of targeting a single domain, our proposed method addresses more general object distributions. Jakab et al. \cite{jakab2020self} used a CycleGAN to transfer unpaired annotations of various objects in form of skeleton edge maps across domains and showed that it applies to the few-shot scenario. We use the same skeleton edge map but use different unsupervised learning techniques to exploit labeled samples more effectively.

\textbf{Unsupervised keypoint localization}
Reconstructing an image from keypoints \cite{jakab2018_unsupervised, zhang2018_unsupervised} and moving keypoints from known or estimated image transformations \cite{thewlis2017_unsupervised} are common approaches in unsupervised keypoint learning. For example, keypoints should be consistent across view transformations in multi-view captures \cite{suwajanakorn2018discovery, rhodin2018unsupervised, Rhodin_2019_CVPR} and have consistent motion transformations in videos \cite{dundar2021unsupervised, siarohin2019_fomm, minderer2019unsupervised, kim2019unsupervised, jakab2020self}. Furthermore, transformations can be created by artificial image deformations in single static images \cite{thewlis2017_unsupervised, zhang2018_unsupervised, jakab2018_unsupervised, lorenz2019_unsupervised}. Another branch is to learn by synthesizing the image without assuming any pre-defined transformation \cite{he2022_autolink, he2021_latentkeypointgan, he2022_ganseg}. Here we adapt the image transformation used in \cite{thewlis2017_unsupervised, lorenz2019_unsupervised} and the image reconstruction technique from \cite{he2022_autolink} into our 3D-aware framework to boost our model performance.

\textbf{Generative image labeling}
Generative models, e.g., GANs \cite{karras2019_stylegan, karras2020_stylegan2} and diffusion models \cite{ho2020_ddpm}, with rich semantic spatial information, can be serve powerful 
priors for solving downstream tasks, e.g., semantic segmentation \cite{zhang2021_datasetgan, tritrong2021_repurposing, yang2022_learning, li2021_semantic, han2022_leveraging}, video object detection \cite{sushko2022_one}, and salient object detection \cite{wu2022_synthetic} from a few annotated examples. 
As these methods require labeling generated images that are prone to artifacts and noise, user annotations might be inaccurate, leading to inaccurate estimated labels \cite{zhang2021_datasetgan}. Also, models trained on generative image-label pairs are limited to the data distribution and biases of the learned image generator \cite{tewari2020_stylerig, or-el2022_stylesdf}. Our approach, however, shows better adaptation to diverse data distributions using a few manual labels.

\section{Method}
\label{sec:method}

\begin{figure*}
    \centering
    \includegraphics[width=\linewidth]{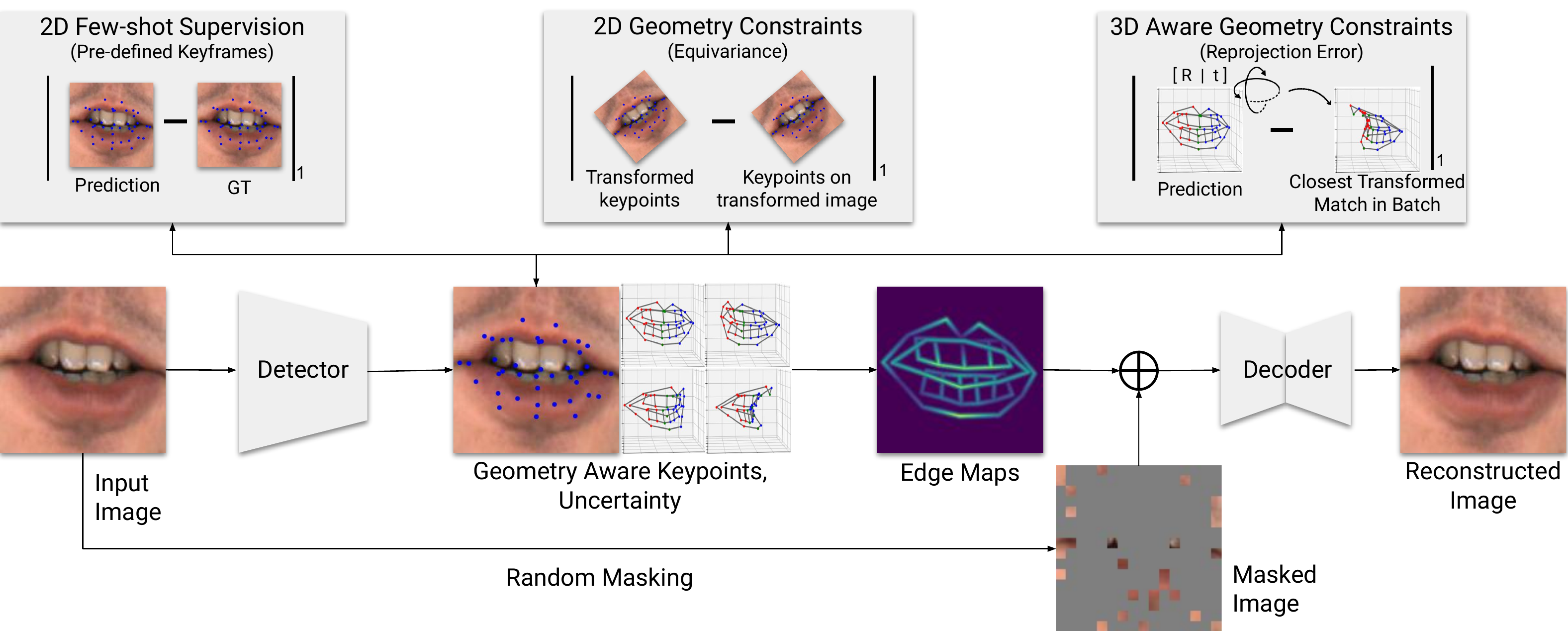}
    \caption{\textbf{Overview}. Given an image, we detect the keypoints and their uncertainty. They are used to generate an edge map, which is concatenated with a randomly masked image to reconstruct the original image. The keypoints are forced to be semantically meaningful by few-shot supervision and consistent by reconstruction. In addition, the 2D and 3D geometric constraints increase the robustness of keypoints.}
    \label{fig:method_overview}
\end{figure*}

Our goal is to learn semantically meaningful and consistent keypoints by using only dozens of annotated examples combined with thousands of unlabeled images.
Our starting point is supervised learning on the dozens of labeled examples. This step is used to define the semantic meaning of keypoints, but alone would horrendously overfit. Hence, it is augmented with self-supervision objectives that encourage meaningful and consistent detection on the additional unlabeled set.
We achieve geometry-aware keypoint localization via a multi-task learning strategy that first detects keypoints with edge and uncertainty maps, and then decodes these maps with randomly masked images to synthesize photo-realistic images, as shown in Figure \ref{fig:method_overview}. 
The edge linkages are provided by the user.
We train both detection and reconstruction stages in an end-to-end manner to ensure that the predicted edges and uncertainty maps derived from keypoints encode the correct semantic object shape definition to synthesize a photo-realistic object. Note that we define the uncertainty differently from previous work \cite{siarohin2019_monkeynet} where it refers to Gaussian shape.

In the following sections, we provide details on the different stages and our few-shot training strategy.

\subsection{Keypoint and Uncertainty Detection}
In this section, we introduce how we obtain the keypoints and uncertainty from the images.
Given an image $\mI\in\R^{H\times W\times 3}$, we use a ResNet with upsamplings \cite{xiao2018_simple-baseline} to predict $K$ heatmaps $\mH_i\in\R^{H\times W}$, and $K$ uncertainty maps $\mV_i\in\R^{H\times W}$, where $i=1...,K$. The 2D keypoints $\vk_i\in\R\in[-1,1]^2$ are generated as the arg-softmax of the heatmap, and the uncertainty $\vv_i$ is calculated as the sum of the map $\mV_i$ weighted by the heatmap, as follows,
\begin{equation}
\begin{aligned}
    &\vk_i = \sum_{\vp}w(\vp)\vp, \quad
    \vv_i = \sum_{\vp}w(\vp)\mV_i(\vp),\\
    &\text{where } w(\vp) = \frac{\exp(\mH_i(\vp))}{\sum_{\vp}\exp(\mH_i(\vp))}.
\end{aligned}
\label{eq:kp_cal}
\end{equation}
Here, $\vp\in[-1,1]^2$ denotes the normalized pixel coordinate.

\subsection{2D Few-shot Supervision} \label{sec:few_shot_supervision}
To leverage the user-provided few-shot 2D keypoint annotations, during each training iteration, we randomly select several image-keypoint pairs, along with a batch of images without any annotations. We concatenate them and penalize deviations of the detected keypoints $\vk'$ from the ground truth keypoints $\vk$ on those with annotations,
\begin{equation}
    \cL_\text{few\_shot}=\frac{1}{|A|}\sum_{i\in A}\|\vk_i-\vk_i'\|_1,
\end{equation}
where $A$ is the set of annotated examples.

\subsection{2D Geometric Constraints}
We force the keypoints to be equivariant to the 2D image transformations, which benefits the robustness as suggested in various unsupervised keypoint detection methods \cite{thewlis2017_unsupervised, jakab2018_unsupervised, lorenz2019_unsupervised}. Let denote $\cT$ as a 2D transformation, which is a combination of affine transformations, flipping and color jitter, and $\vk(\mI)$ as the keypoints detected on image $\mI$. We force the transformed keypoints $\cT(\vk(\mI))$ to be close to the keypoints $\vk(\cT(\mI))$ detected on the transformed image $\cT(\mI)$,
\begin{equation}
    \cL_\text{2d\_geo}=\frac{1}{N}\sum_{i=1}^N\|\vk(\cT(\mI_i))-\cT(\vk(\mI_i))\|_1.
\end{equation}
The keypoints that are out of image boundary after the transformation are ignored. We notice that this equivariance loss significantly harms the performance in the first few iterations during training. Therefore, we linearly increase the range of image transformation based on the number of iterations to stabilize training.
\looseness=-1

\subsection{3D Aware Geometry Constraints and Uplifting}
In the domain of multi-view unsupervised 3D keypoint learning, multi-view consistency is usually used \cite{rhodin2018unsupervised, suwajanakorn2018discovery}. We would like to use 3D consistency for better robustness but face the challenge of not having multiple views to lift to 3D. 

First, to move from a 2D to a 3D keypoint representation, we 
extend
the detector to 
generate $K$ depth maps $\mD_i\in\R^{H\times W}, i=1,...,K$ and calculate the depth as the weighted sum of the depth maps,
\looseness=-1
\begin{equation}
\vd_i = \sum_{\vp}w(\vp)\mD_i(\vp),
\end{equation}
where $w(\vp)$ is the weight defined in Equation~\ref{eq:kp_cal}. The 3D keypoints are defined as the concatenation of the 2D keypoints and the corresponding depths, $\vk^{3D}_i=(\vk_i, \vd_i)$.

Second, we learn these 3D keypoints in the absence of 3D labels and multiple views by exploiting that different instances of the same object are self similar in 3D.
It would be problematic to simply enforce the similarity of all keypoints on different objects as it would force different instances to be exactly the same. To address this issue, we propose to constrain their similarity separately within each part $\cP$.   
Note that the entire object is also included, but as a soft constraint with a lower weight, thus enabling local deformation.
Parts are pre-defined by the user, e.g., the connected components on WFLW face dataset, which are the left/right eye, left/right brow, nose, mouth, and facial contour. 
Specifically, for the keypoints $\vk^{3D}(\mI)$ detected on image $\mI$, we select another set of keypoints $\vk^{3D}(\mI_c)$ detected on image $\mI_c$. For each part $\cP$, i.e., subsets or the whole set of keypoints, we estimate a similarity transformation $\eta$ between the them \cite{umeyama1991_least-squares}, and force the transformed keypoints $\eta(\vk_{\cP}^{3D}(\mI_c))$ 
to be close to the keypoints $\vk_{\cP}^{3D}(\mI)$ of the first example, \looseness=-1
\begin{equation}
    \cL_\text{3d\_geo}=\frac{1}{N}
    \sum_{i=1}^N
    \sum_\cP\frac{1}{|\cP|} \min_{I_c}\|\vk_{\cP}^{3D}(\mI_i)-\eta\left(\vk_{\cP}^{3D}(\mI_c)\right)\|_1.
\label{eq:geo3d}
\end{equation}

To prevent early overfitting, for the first 200 iterations, we randomly pair the parts in the batch. Afterwards, we pair each part with the one in the same batch that minimizes the mean $L_2$ distances as shown in Equation~\ref{eq:geo3d}.
\looseness=-1

\subsection{Geometry-aware Image Reconstruction}
Finally, we exploit that the keypoints should be semantically meaningful enough to reconstruct the original image \cite{he2022_autolink, jakab2018_unsupervised, zhang2018_unsupervised, lorenz2019_unsupervised, jakab2020self}. Specifically, we reconstruct from a largely masked image \cite{he2022_autolink}. Similar to \cite{he2022_autolink}, we introduce an objective to reconstruct the original image from the edge map, with appearance provided by a masked image (90\% of pixels removed). We take the keypoint uncertainty into consideration instead of simply assuming all keypoints are certain as in \cite{he2022_autolink ,jakab2020self}.
Given two keypoints $\vk_i, \vk_j$ defined as ``linked'' by users, we draw a differentiable edge map $\mS_{ij}$, where edge is a Gaussian extended along the line \cite{jakab2020self, he2022_autolink, mihai2021_differentiable}. The values decrease exponentially based on the distance to the line, and are smaller for the uncertain keypoints. Formally, the edge map $\mS_{ij}$ of keypoints $(\vk_i, \vk_j)$ is defined as
\begin{equation}
    \mS_{ij}(\vp) = \exp\left(v_{ij}(\vp)d^2_{ij}(\vp) / \sigma^2\right), 
    \label{eq:edge}
\end{equation}
where $\sigma$ is a learnable parameter controlling the thickness of the edge, $d_{ij}(\vp)$ is the $L_2$ distance between the pixel $\vp$ and the edge drawn by keypoints $\vk_i$ and $\vk_j$, and $v_{ij}(\vp)$ is the uncertainty propagated to pixel $\vp$ along the edge $\vk_i$ to $\vk_j$,
\begin{equation}
\begin{aligned}
    &v_{ij}(\vp) = \left\{
\begin{aligned}
& \text{sigmoid}(\vv_i) &\text{ if } t\leq 0, \\
& \text{sigmoid}((1-t)\vv_i+t\vv_j) &\text{ if } 0<t<1, \\
& \text{sigmoid}(\vv_j) &\text{ if } t\geq 1,
\end{aligned}
\right.\\
    &\text{where}\quad
t = \frac{(\vp-\vk_i)\cdot (\vk_j-\vk_i)}{\|\vk_i-\vk_j\|^2_2}
\end{aligned}
\label{eq:vis}
\end{equation}
is the normalized distance between $\vk_i$ and the projection of $\vp$ onto the edge.
We assign a learnable weight $\alpha$ to edges, which is enforced to be positive by SoftPlus \cite{dugas2000_softplus}. This weight is learned during training and shared across all edges and all object instances in a dataset.
Finally, we take the maximum at each pixel of the heatmaps to obtain the edge map $\mS\in\R^{H\times W}$,
\looseness=-1
\begin{equation}
   \mS(\vp)  = \alpha\max_{ij}\mS_{ij}(\vp).
   \label{eq:max_heatmap}
\end{equation}
Taking the maximum at each pixel avoids the entanglement of the uncertainty and the convolution kernel weights \cite{he2022_autolink}.

The edge map is concatenated with the masked image and fed into a UNet \cite{ronneberger2015_unet} to reconstruct the original image. We minimize the $L_1$ loss and ViT perceptual loss~\cite{tumanyan2022splicing} between the reconstructed images $\mI'$ and the original images $\mI$,
\begin{equation}
    \cL_\text{recon}=\frac{1}{N}\sum_{i=1}^N\|\mI_i-\mI'_i\|_1 + \|\Gamma(\mI_i)-\Gamma(\mI'_i)\|_1,
\end{equation}
where $\Gamma$ is the feature extractor.

\subsection {Coefficients and Few-shot Examples Chosen}
The coefficient for $\cL_\text{few\_shot}, \cL_\text{recon}, \cL_\text{2d\_geo}, \cL_\text{3d\_geo}$ are (1, 1, 1, 0.1), which is decided by the validation sets. There is no other hyperparameters. Note that the validation sets are not used to choose the best model during training but only used to find the loss coefficients, which are shared by all datasets. We believe that using the validation sets for model selection in practice is not available in few-shot learning. The few-shot examples are chosen by the centers of k-means clustering on the features of the 3nd last layer of VGG \cite{simonyan2014_vgg}. More implementation details can be found in the Supplement~\ref{sec:implementation_details}.

\section{Results}
\label{sec:results}

We test our model on 6 diverse datasets, including rigid, soft, and articulated objects, which may contain various appearance and severe occlusion. Notably the mouth interior is extremely challenging due to the large occlusions and had not been attempted before. We compare with three baselines methods: supervised \cite{xiao2018_simple-baseline}, semi-supervised \cite{moskvyak2021_semi-supervised}, and unsupervised \cite{he2022_autolink}. We adapt the latter to few-shot learning 
using the strategy in Section~\ref{sec:few_shot_supervision}. Unless otherwise stated, all results are obtained by 10-shot learning except Tigers where 20 examples are used. 
See Supplement~\ref{sec:datasetgan_comparison} for comparisons with DatasetGAN~\cite{zhang2021_datasetgan}.

\subsection{Datasets}

\begin{figure*}
    \centering
    \includegraphics[width=\linewidth]{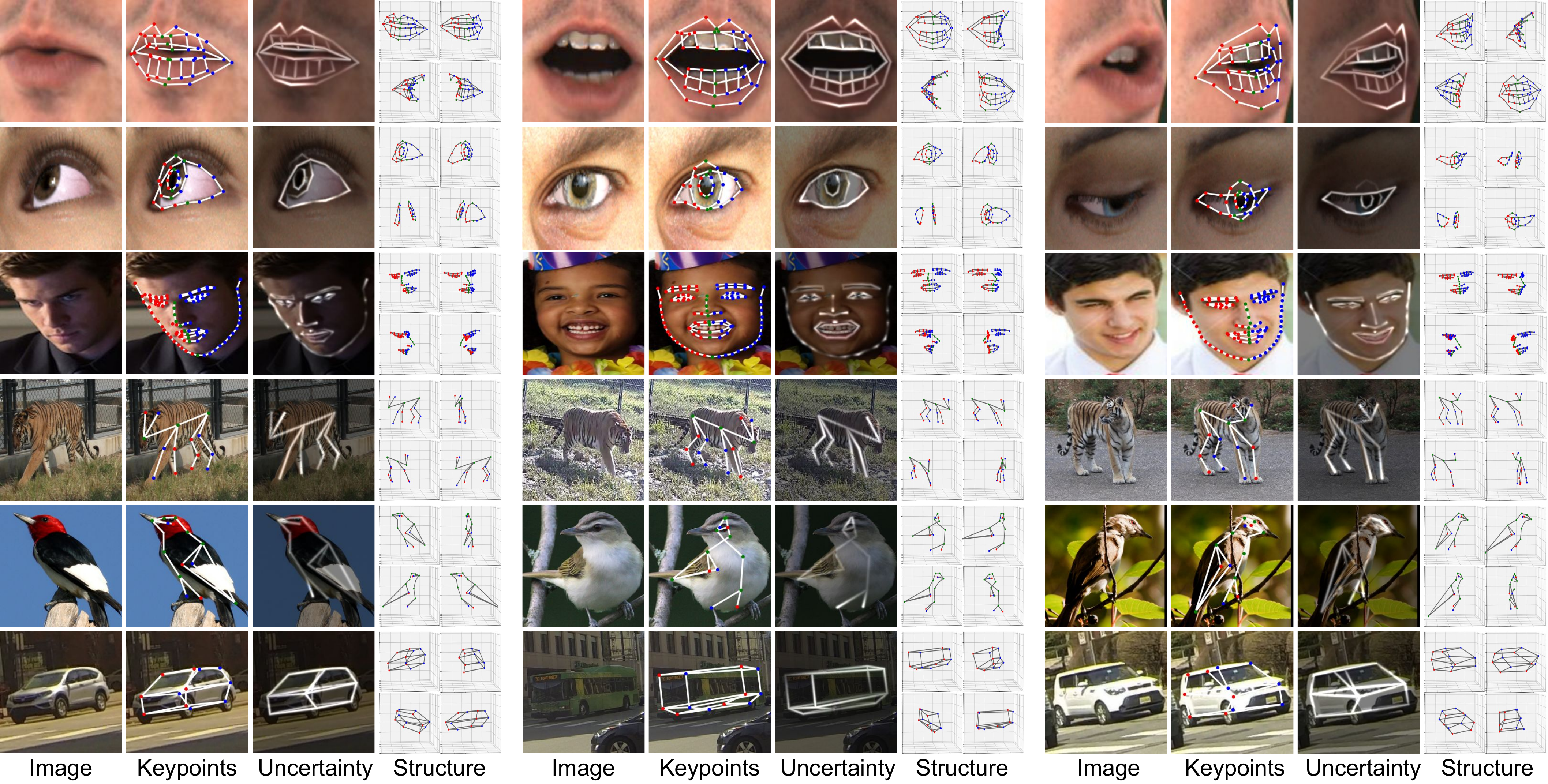}
    \caption{\textbf{Qualitative Results}. With only few shots, the model learns semantically consistent and meaningful keypoints.}
    \label{fig:results}
\end{figure*}

\begin{figure*}
    \centering
    \includegraphics[width=\linewidth]{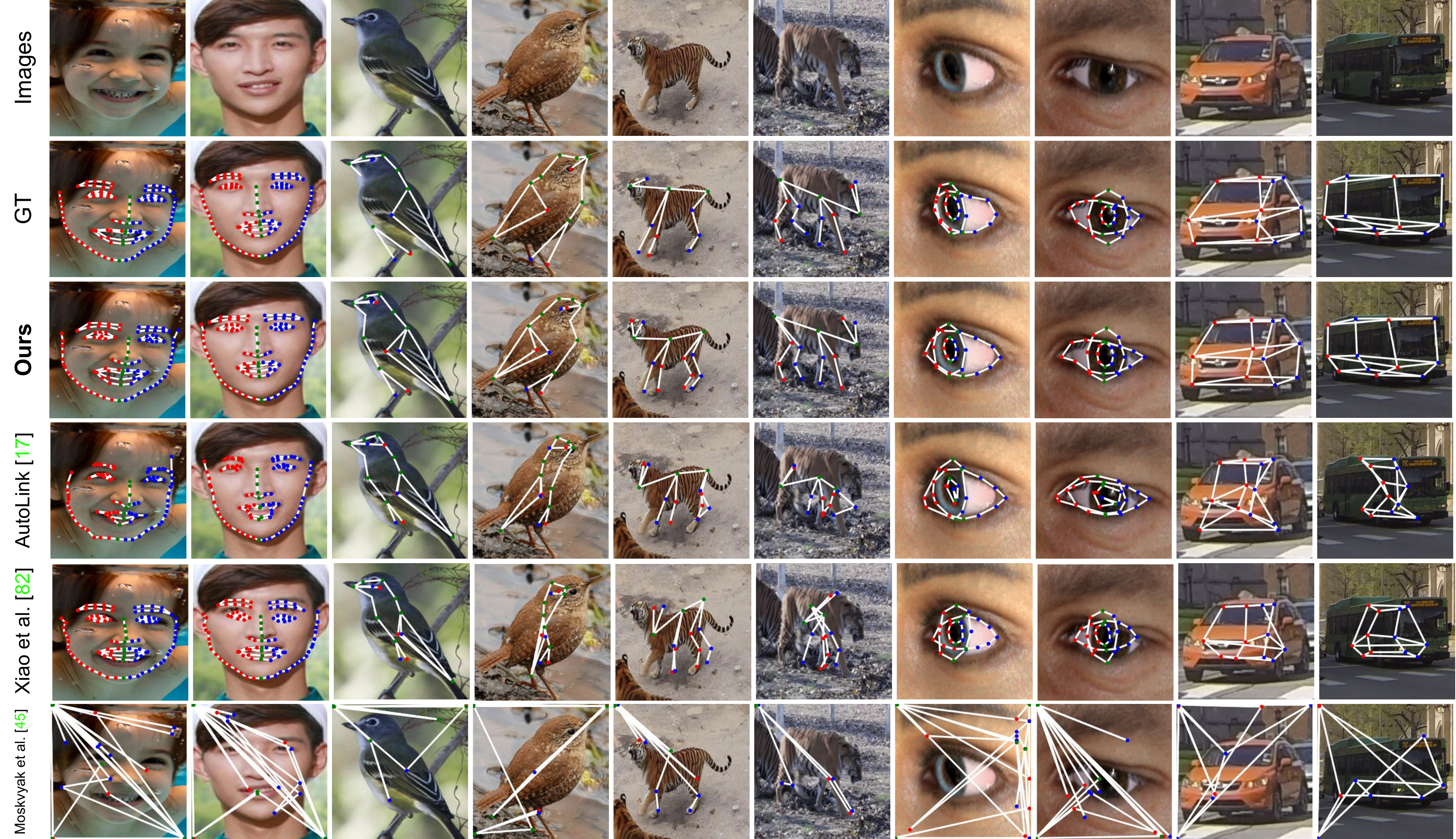}
    \caption{\textbf{Qualitative Comparison}. We qualitatively compare our results with the baselines and the ground truth. The invisible keypoints are not shown in the ground truth if they are not provided. Our performance is significantly better than the other baselines. In the difficult cases, such as tigers and cars, our model still generates good shapes while others fail.}
    \label{fig:comparison}
\end{figure*}

\textbf{MEAD Part0} \cite{wang2020_mead} contains high resolution audio-visual clips of 12 actors. We use the first actor and crop around the mouth. We label 10 images with the lip landmarks from \cite{ferman2022_mdmd} and manual annotations on 4 top and 5 bottom teeth. \looseness=-1

\textbf{SynthesEyes} \cite{wood2015_rendering-eyes} contains 11382 synthesized eye images from 5 male and 5 female subjects. We use the first 4 males and 4 females as our training set and the rest for testing.

\textbf{CUB} \cite{wah2010_cub} contains 200 categories of birds. We use the first 100 for training (5864 images), the next 50 for validation (2958 images), and the last 50 for testing (2966 images). This dataset is to quantitatively evaluate the ability of detecting keypoints on objects of highly various appearance.
    

\textbf{CarFusion} \cite{reddy2018_car-fusion} contains various car images captured in Pittsburgh, PA. We use Fifth Street Part 2 (2794 images) and Part 1 (1597 images) as training and testing sets. \looseness=-1

\textbf{WFLW} \cite{wu2018_wflw} contains 10k faces with a 7500/2500 train/test split. This dataset is more challenging than most face datasets as images have a significant portion of occlusions, make-up, and extreme poses. 

\textbf{ARTW} \cite{li2020_atwr} contains 5159 tiger images captured from multiple wild zoos in unconstrained settings, with a 3610/516/1033 train/val/test split. 

\textbf{SynthesisAI/Faces \cite{synthesisai2022_faces}} contains 10k images of 100 diverse identities with ground truth 3D facial landmarks, including the occluded ones. 
We split the dataset into 7500 images for training and 2500 images for testing.
We use this dataset to quantitatively measure how good the 3D landmarks that our 3D geometry constraint yields are.

\subsection{Qualitative Comparisons}
We show the qualitative results in Figure~\ref{fig:results} and comparisons in Figure~\ref{fig:comparison}. When the number of annotations are limited, the baselines tend to overfit and fail to detect on the difficult cases. For example, only our model successfully detects the keypoints of the bus in the last column of Figure~\ref{fig:comparison}.
\looseness=-1

\subsection{Quantitative Comparisons}
We quantitatively compare our model on 5 benchmarks: WFLW, SynthesEyes, CUB, ATRW, and CarFusion. 

\textbf{Evaluation Metrics} are Normalized $L_2$ error (NME) for SynthesEyes and WFLW, and percentage of correct keypoints (PCK) for CUB, ATRW, and CarFusion. The NME is normalized by distance of eye corners and inter-ocular distances for SynthesEyes and WFLW, respectively. The PCK@0.1 is the ratio of keypoints within a range of 10\% largest bounding box length centered by the GT keypoints. To benefit future research, we also report the other metrics on these datasets in Supplemental~\ref{sec:other_metric}.

Table~\ref{tab:quantitative_comparison} shows that our model significantly outperforms other methods when only 10-50 annotated examples are available. Furthermore, our model is more robust to the number of annotated examples than other methods. For instance, on ATRW, the accuracy of the baselines decrease (58\%/61\%/40\%) if only 50 out of 3610 examples are used, while our performance only decreases 16\%. 
Supplemental~\ref{sec:unsup} provides comparisons with unsupervised methods \cite{jakab2020self, he2022_autolink} on their commonly used datasets: 300W \cite{w300} and H36M \cite{ionescu2014_human3.6m}.

\begin{table}
	\footnotesize
	\begin{center}
		
		\resizebox{1\linewidth}{!}{%
		\begin{tabular}{l| c | c | c | c !{\vrule width 1.5pt} c |c | c }
			\multicolumn{8}{c}{\bfseries NME (\%) on WFLW dataset $\downarrow$}  \\
			\toprule
			\bfseries Method     &  \multicolumn{7}{c}{\bfseries  Training set size}  \\
			&    1    &  10  & 20 &   50   &  5\% & 20\%  & 100\%  \\
			\toprule
			SA \cite{honari2018_semi-supervised} & - & - & - & - & - & \phantom{$\dagger$}6.00$\dagger$ & \textbf{4.39} \\
			Xiao et al. \cite{xiao2018_simple-baseline} & 43.0 & 21.9 & 19.3 & 17.6 & 10.6 & 7.08 & 5.62 \\
			Moskvyak et al. \cite{moskvyak2021_semi-supervised} & 137 & 133 & 76.6 & 21.9 & 10.27 & 6.84 & 6.65 \\
			AutoLink (few) \cite{he2022_autolink} & 14.9 & 13.5 & 13.3 & 11.2 &7.68 & 7.31 &6.35 \\
			3FabRec \cite{browatzki2020_3fabrec} & \phantom{$\dagger$}15.8$\dagger$ & \phantom{$\dagger$}9.66$\dagger$ & - & \phantom{$\dagger$}8.39$\dagger$ &\phantom{$\dagger$}7.68$\dagger$ & \phantom{$\dagger$}6.51$\dagger$ &5.62 \\ 
			\midrule
			ours & \textbf{12.4} & \textbf{9.19} & \textbf{8.62} & \textbf{7.90} & \textbf{6.22} & \textbf{5.61} & 5.38 \\
			\bottomrule
		\end{tabular}	
		}
		\\[6pt]
		\resizebox{1\linewidth}{!}{%
		\begin{tabular}{l| c | c | c | c !{\vrule width 1.5pt} c |c | c }
			\multicolumn{8}{c}{\bfseries NME (\%) on SynthesEyes dataset $\downarrow$}  \\
			\toprule
			\bfseries Method     &  \multicolumn{7}{c}{\bfseries  Training set size}  \\
			&    1    &  10  & 20 &   50   &  5\% & 20\%  & 100\%  \\
			\toprule
			Xiao et al. \cite{xiao2018_simple-baseline} & 25.1 & 15.9 & 10.6 & 8.11 & 4.07 & \textbf{3.12} & 2.65 \\
			Moskvyak et al. \cite{moskvyak2021_semi-supervised} & 91.1 & 86.8 & 45.8 & 18.5 & 4.52 & 3.24 & \textbf{2.49} \\
			AutoLink (few) \cite{he2022_autolink} & 26.4 & 14.2 & 8.78 & 7.80 & 4.28 & 3.32 & 2.86 \\
			\midrule
			ours & \textbf{24.0} & \textbf{6.93} & \textbf{6.83} & \textbf{5.50} & \textbf{3.69} & 3.13 & 2.96 \\
			\bottomrule
		\end{tabular}	
		}
		\\[6pt]
		\resizebox{1\linewidth}{!}{%
		\begin{tabular}{l| c | c | c | c !{\vrule width 1.5pt} c |c | c }
			\multicolumn{8}{c}{\bfseries PCK@0.1 (\%) on CUB-200-2011 dataset $\uparrow$}  \\
			\toprule
			\bfseries Method     &  \multicolumn{7}{c}{\bfseries  Training set size}  \\
			&    1    &  10  & 20 &   50   &  5\% & 20\%  & 100\%  \\
			\toprule
			Xiao et al. \cite{xiao2018_simple-baseline} & 6.01 & 31.2 & 35.3 & 42.8 & 60.5 & 73.9 & 90.5 \\
			Moskvyak et al. \cite{moskvyak2021_semi-supervised} & 7.38 & 20.9 & 28.3 & 63.2 & \phantom{$\dagger$}\textbf{91.1}$\dagger$ & \phantom{$\dagger$}\textbf{92.4}$\dagger$ & \textbf{93.8} \\
			AutoLink (few) \cite{he2022_autolink} & \textbf{26.2} & 35.1 & 41.2 & 51.8 & 67.2 & 75.9 &87.6 \\
			\midrule
			ours & 16.3 & \textbf{70.7} & \textbf{73.1} & \textbf{75.1} & 84.2 & 88.3 & 90.1 \\
			\bottomrule
		\end{tabular}	
		}
		\\[6pt]
		\resizebox{1\linewidth}{!}{%
		\begin{tabular}{l| c | c | c | c !{\vrule width 1.5pt} c |c | c }
			\multicolumn{8}{c}{\bfseries PCK@0.1 (\%) on ATRW dataset $\uparrow$}  \\
			\toprule
			\bfseries Method     &  \multicolumn{7}{c}{\bfseries  Training set size}  \\
			&    1    &  10  & 20 &   50   &  5\% & 20\%  & 100\%  \\
			\toprule
			Xiao et al. \cite{xiao2018_simple-baseline} & 13.2 & 21.5 & 22.5 & 23.9 & 51.6 & 86.1 & 96.1 \\
			Moskvyak et al. \cite{moskvyak2021_semi-supervised} & 3.38 & 17.9 & 27.5 & 57.1 & \phantom{$\dagger$}92.6$\dagger$ & \phantom{$\dagger$}94.5$\dagger$ & 95.3 \\
			AutoLink (few) \cite{he2022_autolink} & \textbf{20.8} & 22.1 & 33.8 & 37.4 & 77.8 & 89.4 & \textbf{97.1} \\
			\midrule
			ours & 14.4 & \textbf{36.3} & \textbf{78.8} & \textbf{81.4} & \textbf{92.7} & \textbf{96.5} & 96.9 \\
			\bottomrule
		\end{tabular}	
		}
		\\[6pt]
		\resizebox{1\linewidth}{!}{%
		\begin{tabular}{l| c | c | c | c !{\vrule width 1.5pt} c |c | c }
			\multicolumn{8}{c}{\bfseries PCK@0.1 (\%) on CarFusion dataset $\uparrow$}  \\
			\toprule
			\bfseries Method     &  \multicolumn{7}{c}{\bfseries  Training set size}  \\
			&    1    &  10  & 20 &   50   &  5\% & 20\%  & 100\%  \\
			\toprule
			Xiao et al. \cite{xiao2018_simple-baseline} & 11.7 & 23.4 & 30.0 & 37.5 & 51.1 & 77.9 & 89.7 \\
			Moskvyak et al. \cite{moskvyak2021_semi-supervised} & 3.84 & 5.38 & 22.4 & 34.7 & 64.7 & 87.3 & 92.5 \\
			AutoLink (few) \cite{he2022_autolink} & \textbf{14.5} & 31.9 & 42.3 & 62.2 & 69.9 & 83.5 &90.8 \\
			\midrule
			ours & 13.8 & \textbf{66.7} & \textbf{68.8} & \textbf{75.5} & \textbf{81.6} & \textbf{93.5} & \textbf{93.8} \\
			\bottomrule
		\end{tabular}
		}
		
	\end{center}
	\vspace{-0.5cm}
	\caption{\textbf{Quantitative Comparison}. In the few-shot scenario, where only 10-50 annotated examples are available, our model significantly outperforms the baselines. The sign $\dagger$ means the number is reported in another set of examples used in their papers.}
	\label{tab:quantitative_comparison}
\end{table}

\begin{table}
	\footnotesize
	\begin{center}
	\resizebox{1\linewidth}{!}{%
		\begin{tabular}{l| c | c | c | c !{\vrule width 1.5pt} c |c | c}
			\toprule
			\bfseries Methods     &  \multicolumn{7}{c}{\bfseries  Training set size}  \\
			&    1    &  10  & 20 &   50   &  5\% & 20\%  & 100\%  \\
			\toprule
			Xiao et al. \cite{xiao2018_simple-baseline} & 36.5 & 22.5 & 22.1 & 17.8 & 8.33 & \textbf{5.37} & \textbf{4.19} \\
			\midrule
			Full & \textbf{15.6} & \textbf{11.2} & \textbf{9.25} & \textbf{8.46} & \textbf{7.07} & 6.44 & 5.99 \\
			\bottomrule
		\end{tabular}	
		}
	\end{center}
	\vspace{-0.5cm}
	\caption{\textbf{NME(\%) on SynthesisAI 3D Faces.} We compare our approach with a supervised baseline. Note that the latter uses ground truth 3D keypoints, while ours only needs 2D keypoints.}	
	\label{tab:face3d}
\end{table}

\section{Analysis}
\label{sec:analysis}

We test the effects and the necessity of our designed modules. Quantitative comparisons are implemented on WFLW and ATRW, with the same settings and the same few-shot examples as in Section~\ref{sec:results}.

\textbf{Image Reconstruction}. Table \ref{tab:ablation_wflw} shows that without the image Reconstruction constraint, the error increases dramatically for low numbers of annotated examples. It is believed that reconstructing from edge maps aligns the edges with the object edges \cite{he2022_autolink}, which stabilize the keypoints in the few-shot learning. Note that with more than 20\% annotated examples, the image reconstruction harms the performance, probably because the reconstruction task drifts the gradient from accurate keypoint supervision gradient.

\textbf{Geometric Constraint}. While both 2D and 3D geometric constraints increase accuracy, their insights differ slightly. The 2D constraint is at the image level, which is accurate. It works as data augmentation on learned keypoints, which is expected to increase the robustness. Interestingly, the variant without 2D geometric constraint has better performance on Tigers (10-shots). We believe it is due to the symmetric shape of the tigers, as explained in the limitations. On the other hand, the 3D constraint is an approximation and enforces similarity between two different examples. This constraint is expected to prevent the model from generating extreme outliers. Its effects on articulated objects, e.g., tigers, are not as significant as they are on soft and rigid objects, e.g., faces. 

\begin{figure}
    \centering
    \includegraphics[width=\linewidth]{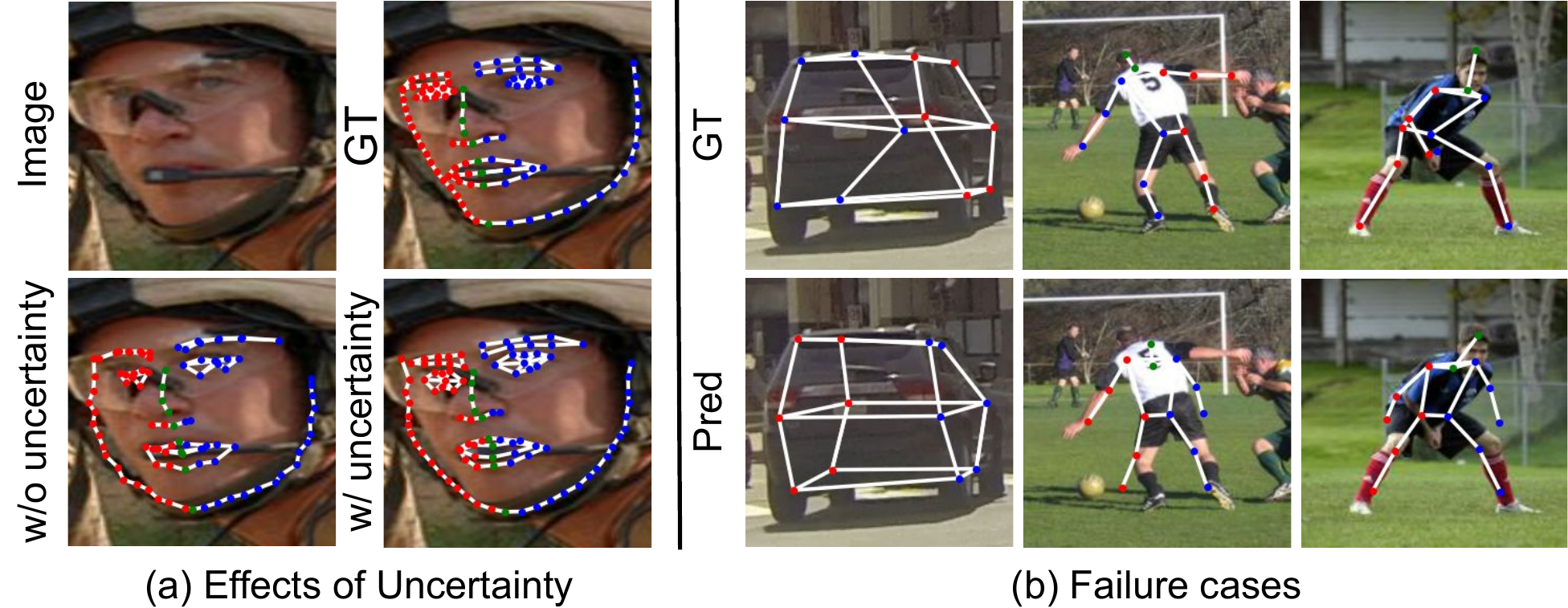}
    \caption{\textbf{(a) Effects of Uncertainty}. Modeling uncertainty improves keypoint localization on objects with occlusion (see for instance the nose and eyebrows). \textbf{(b) Failure Cases}. The model fails if the objects are highly symmetric or articulated.}
    \label{fig:limitation}
\end{figure}

\textbf{Uncertainty}. Uncertainty models occluded and ambiguous edges in image reconstruction by making the uncertain edges lighter. Without it, the model may be confused in the reconstruction stage about whether occluded edges should be drawn. As a result, it harms performance. Figure~\ref{fig:limitation}(a) shows an example of a largely occluded face. The model without uncertainty gives an average face. On ATRW, the difference is less obvious as the keypoints are visible in most cases.\looseness=-1

\begin{figure}
    \centering
    \includegraphics[width=\linewidth]{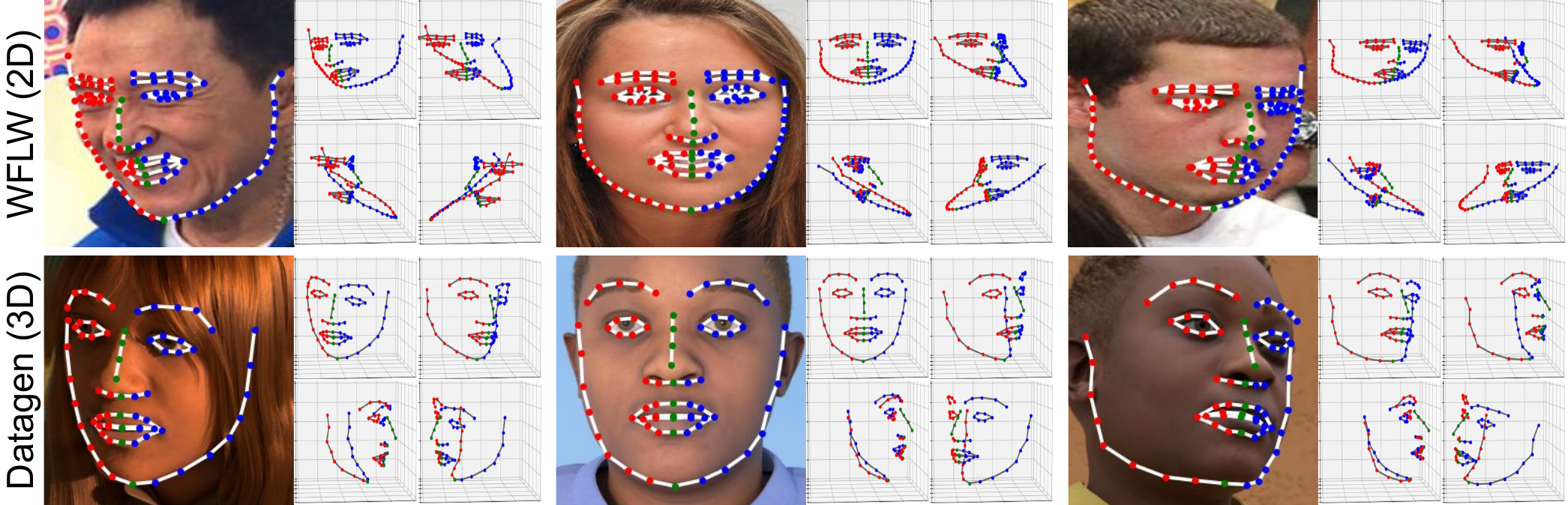}
    \caption{\textbf{Detection on 2D and 3D-projected Face Landmarks}. If the model is trained on the 2D face dataset WFLW, the learned jaw landmarks are not 3D consistent. However, if it is trained on the synthetic 3D-projected landmarks, the jaw contours are 3D consistent, though the projection is still on the 2D facial contour.}
    \label{fig:face3d}
\end{figure}

\textbf{How good are the 3D keypoints?} We train on the SynthesisAI/Faces \cite{synthesisai2022_faces} using only 2D landmarks and compare the 3D landmarks quality with the supervised baseline \cite{xiao2018_simple-baseline}, where depth is additionally learned. We evaluate the learned 3D landmarks qualitatively in Figure~\ref{fig:face3d} and quantitatively in Table~\ref{tab:face3d}. Figure~\ref{fig:face3d} also shows the difference between the jaw landmarks learned from a 2D and 3D-projected facial landmark dataset.
The signal provided by the latter gives a better 3D jaw shape since the few-shot landmarks do not follow visible image boundaries as those in the 2D facial dataset.\looseness=-1

\textbf{Few-shot Example Selection}. We tested replacing the KMeans with random selection.
Table~\ref{tab:ablation_wflw} reports the average error over 3 runs.
When only dozens of annotated examples are available, it is important to pick the most representative ones, especially for articulated objects, such as ATRW tigers.

\begin{table}
	\footnotesize
	\begin{center}
	    \resizebox{1\linewidth}{!}{%
		\begin{tabular}{l| c | c | c | c !{\vrule width 1.5pt} c |c | c}
		\multicolumn{8}{c}{\bfseries NME (\%) on WFLW dataset $\downarrow$}  \\
			\toprule
			\bfseries Variants     &  \multicolumn{7}{c}{\bfseries  Training set size}  \\
			&    1    &  10  & 20 &   50   &  5\% & 20\%  & 100\%  \\
			\toprule
			- image reconstruction & 27.4 & 24.8 & 20.2 & 12.4 & 6.49 & \textbf{5.32} & \textbf{4.69} \\ 
			- 2D geometry & 16.7 & 14.2 & 13.7 & 12.6 & 9.24 & 7.81 & 6.83 \\ 
			- 3D geometry & 15.5 & 12.4 & 12.1 & 11.3 & 7.67 & 6.47 & 6.37 \\ 
			- uncertainty & 13.5 & 12.6 & 11.8 & 10.4 & 7.28 & 6.75 & 6.42 \\ 
			- kmeans selection & 24.2 & 14.5 & 12.6 & 10.8 & 6.81 & 5.54 & - \\ 
			\midrule
			Full & \textbf{12.4} & \textbf{9.19} & \textbf{8.62} & \textbf{7.90} & \textbf{6.22} & 5.61 & 5.38 \\
			\bottomrule
		\end{tabular}	
		}
		\\[6pt]
		\resizebox{1\linewidth}{!}{%
		\begin{tabular}{l| c | c | c | c !{\vrule width 1.5pt} c |c | c}
		\multicolumn{8}{c}{\bfseries PCK@0.1 (\%) on ATRW dataset $\uparrow$}  \\
			\toprule
			\bfseries Variants     &  \multicolumn{7}{c}{\bfseries  Training set size}  \\
			&    1    &  10  & 20 &   50   &  5\% & 20\%  & 100\%  \\
			\toprule
			- image reconstruction & 20.2 & 22.1 & 22.8 & 56.5 & 91.0 & \textbf{96.8} & \textbf{97.6} \\ 
			- 2D geometry & 24.7 & \textbf{40.6} & 39.3 & 39.6 & 87.4 & 93.7 & 96.1 \\ 
			- 3D geometry & 16.7 & 36.2 & 75.8 & 81.2 & 92.1 & 96.1 & 96.4 \\ 
			- uncertainty & 20.8 & 31.9 & 78.7 & 81.2 & 92.3 & 95.9 & 95.4 \\ 
			- kmeans selection & 8.2 & 10.1 & 15.1 & 40.7 & 90.9 & 96.1 & - \\ 
			\midrule
			Full & 14.4 & 36.3 & \textbf{78.8} & \textbf{81.4} & \textbf{92.7} & 96.5 & 96.9 \\
			\bottomrule
		\end{tabular}	
		}
	\end{center}
	\vspace{-0.5cm}
	\caption{\textbf{Ablation Tests on WFLW and ATRW.} Each of our design choices plays an important role in the few-shot scenario.}	
	\label{tab:ablation_wflw}
\end{table}

\textbf{Limitations.} \label{sec:limitation}
There are two limitations especially when the annotated dataset is very small. First, if the object's keypoints are highly symmetric, there may be left-right or front-back ambiguity. Second, if the object is highly articulated, e.g., humans in LSP dataset \cite{johnson2010_lsp}, the estimated keypoints are less accurate. Figure~\ref{fig:limitation}(b) illustrates both cases.
However, neither problem is observed when we increase the dataset to hundreds of annotations (instead of dozens).

\section{Conclusion \& Discussion}
\label{sec:conclusion}

We presented a few-shot keypoint localization method that is formed by combining keypoint detection with uncertainty, 2D/3D geometric constraints, and image reconstruction. These components prevent the detector from overfitting to the few-shot examples and utilize unlabelled images. Our experiment results demonstrate that, with only dozens of annotations, our model works on various datasets, including rigid, soft, articulated objects, and even the very difficult mouth interior which has not been tried before. In the few-shot scenario, our model significantly outperforms the baselines and works on those datasets where others fail with 10 or 20 annotated examples.
It opens the path for conditional generative modeling and image editing with a few annotated examples.
Our future work will focus on leveraging 3D-aware image synthesis for better generalization to extreme poses, solving the symmetry problem, and testing on broader object categories.
\looseness=-1


{\small
\bibliographystyle{ieee_fullname}
\bibliography{00.main}
}

\clearpage
\appendix
\section {Implementation details} \label{sec:implementation_details}
\paragraph{Optimization} All images are resized to $128\times 128$. We use the Adam optimizer \cite{kingma2014_adam} with a learning rate of $10^{-4}$ with $\beta_1=0.9$, $\beta_2=0.99$. The batch size is 16 for unlabeled images and $\min(16, \text{n\_examples})$ for few-shot examples. We train for 20k iterations. The gradients are stopped in the similarity transformation estimation in the 3D geometric constraint so that the shape instead of transformation itself would be optimized. The ViT perceptual loss is based on the last attention keys and the the global context vector. For the reconstruction, we divide the image into $16\times 16$ patches and randomly mask 90\%. The random seeds for all packages are fixed to 0.

\paragraph{Formula of $\vd_{ij}$} The distance $\vd_{ij}$ from a pixel $\vp$ to an edge drawn by keypoints $\vk_i$ and $\vk_j$ in Equation~\ref{eq:edge} is
\begin{equation}
\begin{aligned}
    &d_{ij}(\vp) = \left\{
\begin{aligned}
& \|\vp-\vk_i\|_2 &\text{ if } t\leq 0, \\
& \|\vp-((1-t)\vk_i + t\vk_j)\|_2 &\text{ if } 0<t<1, \\
& \|\vp-\vk_j\|_2 &\text{ if } t\geq 1,
\end{aligned}
\right.\\
    &\text{where}\quad
t = \frac{(\vp-\vk_i)\cdot (\vk_j-\vk_i)}{\|\vk_i-\vk_j\|^2_2}
\end{aligned}
\label{eq:dis_pt2line}
\end{equation}
is the normalized distance between $\vk_i$ and the projected $\vp$ onto the edge as in \ref{eq:vis}.

\paragraph{Formulation of $\sigma$ and $\alpha$} The thickness $\sigma$ in Equation~\ref{eq:edge} is formulated as
\begin{equation}
    \sigma^2 = 1 / 1000\exp\theta, 
\end{equation}
where $\theta$ is learnable and initialized to 1. 

The edge map weight $\alpha$ in Equation~\ref{eq:max_heatmap} is formulated as
\begin{equation}
    \alpha = \text{SoftPlus}(\gamma), 
\end{equation}
where $\gamma$ is learnable and initialized to -4.

\paragraph{2D geometric constraint} The image transformation in 2D geometric constraint is a combination of random rotation ($-60\sim60$), translation ($-10\%\sim10\%$), scaling ($0.9\sim1$), flipping (p=0.5), and color jitter (brightness, contrast, saturation, hue from -50\% to 50\%). The augmentation ranges are multiplied by 0 and 1 at iteration 0 and 20k, respectively. Note that this coefficient increases linearly from 0 to 1 during training.

\paragraph{Facial landmark smoothing} On WFLW and SynthesisAI/Faces, we encourage the cosine of the angle of the two neighboring landmarks to be 0. This penalty has weight 0.02, and is only for jaws and noses.

\section{Comparison with DatasetGAN} \label{sec:datasetgan_comparison}
We use the pre-trained StyleGAN generator on FFHQ \cite{karras2019_stylegan} of resolution $256\times 256$. We sample 10000 images and choose 10 images by the centers of k-means clustering on the features of the 3rd last layer of VGG \cite{simonyan2014_vgg}. 
The keypoints are annotated by DLIB \cite{dlib09}, which is originally used for FFHQ alignment.

To train our model, we use the first 60000 images for training and the last 10000 images for testing. To make a fair comparison, we train our model in two different sets of few-shot examples: 1) pick from the dataset as in the main paper; 2) use the generated examples, akin to DatasetGAN. 

The results are summarized in Table~\ref{tab:comparison_datasetgan}. Our model outperforms DatasetGAN in all different number of annotated examples. Note that their StyleGAN model is trained on all 70000 images in the dataset while our unlabeled dataset only contains the first 60000 images. 

We remark that our model trained on the generated examples is not as good as those trained on real images. This demonstrates that the artifacts and noise in the generated images have a significant impact on the keypoint localization.
\begin{table}
	\footnotesize
	\begin{center}
	
		\resizebox{1\linewidth}{!}{%
		\begin{tabular}{l| c | c | c | c }
			\multicolumn{5}{c}{\bfseries NME (\%) on FFHQ dataset $\downarrow$}  \\
			\toprule
			\bfseries Method     &  \multicolumn{3}{c}{\bfseries  Training set size}  \\
			&    1    &  10  & 20 &   50\\
			\toprule
			DatasetGAN \cite{zhang2021_datasetgan} & 18.4 & 8.04 & 7.24 & 5.50  \\
			\midrule
			ours (trained with generated examples) & 14.2 & 7.08 & 6.37 & 5.22 \\
			ours & \textbf{11.3} & \textbf{5.87} & \textbf{5.89} & \textbf{4.59} \\
			\bottomrule
		\end{tabular}	
		}
		
	\end{center}
	\vspace{-0.5cm}
	\caption{\textbf{Quantitative Comparison with DatasetGAN on FFHQ dataset.} }	
	\label{tab:comparison_datasetgan}
\end{table}

\begin{figure}
    \centering
    \includegraphics[width=\linewidth]{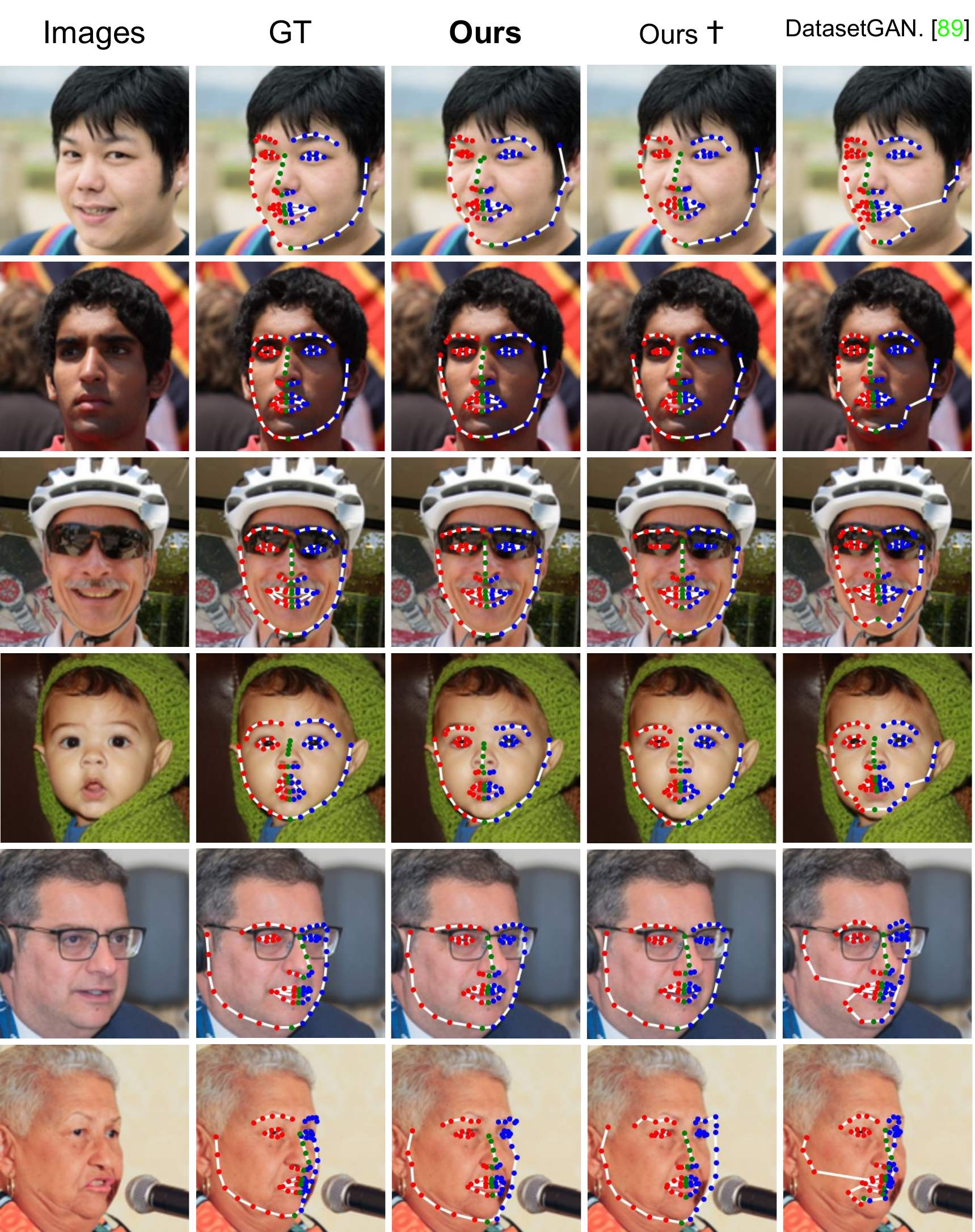}
    \caption{\textbf{Comparison with DatasetGAN}. Our model generates better shapes that are closer to the ground truth than DatasetGAN. The third row and fourth row with symbol $\dagger$ are obtained by training on the real annotated images and the synthetic annotated images, respectively.}
    \label{fig:comparison_datasetgan}
\end{figure}

\section{Comparison on LSP Dataset}
For completeness, we also show the results on LSP~\cite{johnson2010_lsp} dataset in Table~\ref{tab:limitation_lsp}. The metric and train/test split follow \cite{moskvyak2021_semi-supervised}. Our accuracy in the few-shot scenario is significantly higher than all the other baseline methods.
\begin{table}
	\footnotesize
	\begin{center}
	
		\resizebox{1\linewidth}{!}{%
		\begin{tabular}{l| c | c | c | c !{\vrule width 1.5pt} c |c | c }
			\multicolumn{8}{c}{\bfseries PCK@0.1 (\%) on LSP dataset}  \\
			\toprule
			\bfseries Method     &  \multicolumn{7}{c}{\bfseries  Training set size}  \\
			&    1    &  10  & 20 &   50   &  5\% & 20\%  & 100\%  \\
			\toprule
			Xiao et al. \cite{xiao2018_simple-baseline} & \textbf{15.3} & 22.1 & 23.9 & 26.5 & 24.0 & 47.8 & 71.1 \\
			AutoLink \cite{he2022_autolink} & 19.7 & 22.3 & 26.9 & 35.4 & 35.3 & 50.0 &75.0 \\
			Moskvyak et al. \cite{moskvyak2021_semi-supervised} & 3.89 & 7.13 & 15.1 & 27.7 & \textbf{67.0} & \textbf{71.9} & 74.3 \\
			\midrule
			ours & 14.2 & \textbf{49.8} & \textbf{53.2} & \textbf{58.1} & 64.5 & 70.4 & \textbf{77.9} \\
			\midrule
			\bottomrule
		\end{tabular}	
		}
		
	\end{center}
	\vspace{-0.5cm}
	\caption{\textbf{Quantitative Comparison with Baselines on LSP.}}	
	\label{tab:limitation_lsp}
\end{table}

\section{Comparison with Unsupervised Methods} \label{sec:unsup}
In Table~\ref{tab:quantitative_comparison_unsup}, we compare our method with state-of-the-art unsupervised methods on their commonly used datasets, i.e., 300W \cite{w300} and h36m \cite{ionescu2014_human3.6m}. We follow the evaluation protocol described in \cite{jakab2020self} to perform our comparisons. In particular, to match the training and evaluation scheme of unsupervised approaches \cite{jakab2020self, he2022_autolink, zhang2018_unsupervised} on h36m dataset, where the left and right side of the object is ambiguous during training, we flip all few-shot skeletons to facing front. At inference, we choose the correct left and right side by simply flipping the skeleton back to the correct orientation\cite{zhang2018_unsupervised} if needed.

Our method outperforms the state of the art on both datasets in the [10-50]-shot scenario.
For evaluating AutoLink \cite{he2022_autolink}, besides adding few-shot supervision as described in Section~\ref{sec:few_shot_supervision}, we also follow the traditional way in unsupervised learning \cite{thewlis2017_unsupervised, jakab2020self, he2022_autolink, zhang2018_unsupervised}. Specifically, we fit a linear regression model from the unsupervised keypoints to the annotated keypoints. However, linear regression is not a data efficient approach. It requires more labels than the AutoLink (few) baseline and 
100x more labels than our method, even though we chose the best $L_2$ regularization coefficient in [0, 10] and the optimal number of keypoints in [4,32] by grid search (10-fold cross-validation).
Note that Jakab et al. \cite{jakab2020self} use unpaired annotations which are beneficial for transferring semantics across domains (e.g., sim2real) but cannot exploit to the full extent when image/pose pairs are available. What we claim as a contribution is a novel formulation that includes labeled examples and adds 3D and visibility constraints, altogether leading to substantial improvements.
\begin{table}
	\footnotesize
	\begin{center}
	
		\resizebox{1\linewidth}{!}{%
		\begin{tabular}{l| c | c | c | c !{\vrule width 1.5pt} c |c | c }
			\multicolumn{8}{c}{\bfseries NME (\%) on H36M dataset $\downarrow$}  \\
			\toprule
			\bfseries Method     &  \multicolumn{7}{c}{\bfseries  Training set size}  \\
			&    1    &  10  & 20 &   50   &  500 & 5000 & 100\%  \\
			\toprule
			Xiao et al. \cite{xiao2018_simple-baseline} & 11.8 & 5.53 & 5.33 & 4.71 & 3.44 & 2.35 & \textbf{2.04} \\
			Moskvyak et al. \cite{moskvyak2021_semi-supervised} & 119 & 104 & 51.3 & 18.2 & 2.99 & \textbf{2.30} & 2.06 \\
			AutoLink (reg) \cite{he2022_autolink} & 18.6 & 10.5 & 9.16 & 7.94 & 3.85 & 3.00 & 2.74 \\
			AutoLink (few) \cite{he2022_autolink} & 9.62 & 5.49 & 4.09 & 3.76 & 3.07 & 2.62 & 2.55 \\
			Jakab et al. \cite{jakab2020self} & - & - & - & 4.05 & 3.30 & 2.92 & 2.73 \\
			\midrule
			ours & \textbf{7.53} & \textbf{4.21} & \textbf{3.67} & \textbf{3.14} & \textbf{2.84} & 2.57 & 2.58 \\
			\bottomrule
		\end{tabular}	
		}
		\\[6pt]
		\resizebox{1\linewidth}{!}{%
		\begin{tabular}{l| c | c | c | c !{\vrule width 1.5pt} c |c | c }
			\multicolumn{8}{c}{\bfseries NME (\%) on 300W dataset $\downarrow$}  \\
			\toprule
			\bfseries Method     &  \multicolumn{7}{c}{\bfseries  Training set size}  \\
			&    1    &  10  & 20 &   50   &  500 & 5000 & 100\%  \\
			\toprule
			Xiao et al. \cite{xiao2018_simple-baseline} & 20.2 & 15.0 & 12.8 & 11.0 & 6.37 & - & \textbf{4.74} \\
			Moskvyak et al. \cite{moskvyak2021_semi-supervised} & 85.8 & 83.3 & 46.3 & 25.1 & 8.15 & - & 4.84 \\
			AutoLink (reg) \cite{he2022_autolink} & 25.4 & 10.7 & 9.90 & 8.31 & 6.07 & - & 5.63 \\
			AutoLink (few) \cite{he2022_autolink} & 15.2 & 9.85 & 9.82 & 8.61 & 6.70 & - & 5.25 \\
			Jakab et al. \cite{jakab2020self} & - & - & - & 8.92 & 8.91 & - & 8.67 \\
			\midrule
			ours & \textbf{9.45} & \textbf{7.61} & \textbf{6.82} & \textbf{6.25} & \textbf{5.58} & - & 4.96 \\
			\bottomrule
		\end{tabular}	
		}
		\caption{\textbf{Quantitative Comparison with Unsupervised Methods on 300W and H36M datasets.}}
	\label{tab:quantitative_comparison_unsup}
	\end{center}
	\vspace{-0.5cm}	
\end{table}

\section{Additional Analysis of Jaw Landmarks}
We notice that the top two jaw landmarks tend to be farther from their neighbors than the other jaw landmarks. We believe that it is due to the image reconstruction. The model tends to model the foreheads with the top two jaw landmarks so that the head structure is more clear to the model. As a result, the reconstruction has better quality.

\section{More Quantitative Results} \label{sec:other_metric}
In Table~\ref{tab:quantitative_comparison} in Section~\ref{sec:results}, we report NME on WFLW and SynthesEyes, and PCK on CUB, ATRW, CarFusion. To promote future research, we also report additional metrics in Table~\ref{tab:quantitative_comparison_other}, namely PCK on WFLW and SynthesEyes, and NME on CUB, ATRW, and CarFusion.
\begin{table}
	\footnotesize
	\begin{center}
		
		\resizebox{1\linewidth}{!}{%
		\begin{tabular}{l| c | c | c | c !{\vrule width 1.5pt} c |c | c }
			\multicolumn{8}{c}{\bfseries PCK@0.1 (\%) on WFLW dataset $\uparrow$}  \\
			\toprule
			\bfseries Method     &  \multicolumn{7}{c}{\bfseries  Training set size}  \\
			&    1    &  10  & 20 &   50   &  5\% & 20\%  & 100\%  \\
			\toprule
			Xiao et al. \cite{xiao2018_simple-baseline} & 4.73 & 32.1 & 36.3 & 44.4 & 68.2 & 81.5 & 87.0 \\
			Moskvyak et al. \cite{moskvyak2021_semi-supervised} & 4.25 & 4.85 & 17.7 & 49.6 & 58.1 & 84.9 & 85.4 \\
			AutoLink (few) \cite{he2022_autolink} & 47.7 & 53.4 & 55.0 & 63.1 & 75.5 & 80.2 & 83.7 \\
			\midrule
			ours & \textbf{58.3} & \textbf{71.0} & \textbf{73.9} & \textbf{77.3} & \textbf{84.1} & \textbf{86.7} & \textbf{87.5} \\
			\bottomrule
		\end{tabular}	
		}
		\\[6pt]
		\resizebox{1\linewidth}{!}{%
		\begin{tabular}{l| c | c | c | c !{\vrule width 1.5pt} c |c | c }
			\multicolumn{8}{c}{\bfseries PCK@0.1 (\%) on SynthesEyes dataset $\uparrow$}  \\
			\toprule
			\bfseries Method     &  \multicolumn{7}{c}{\bfseries  Training set size}  \\
			&    1    &  10  & 20 &   50   &  5\% & 20\%  & 100\%  \\
			\toprule
			Xiao et al. \cite{xiao2018_simple-baseline} & 12.5 & 37.7 & 61.4 & 75.4 & 96.1 & 99.0 & \textbf{99.7} \\
			Moskvyak et al. \cite{moskvyak2021_semi-supervised} & 3.09 & 9.74 & 25.2 & 38.8 & 95.0 & \textbf{99.2} & 99.4 \\
			AutoLink (few) \cite{he2022_autolink} & 25.8 & 49.9 & 73.6 & 82.4 & 95.6 & 98.7 & \textbf{99.7} \\
			\midrule
			ours & \textbf{32.9} & \textbf{77.6} & \textbf{79.4} & \textbf{89.9} & \textbf{97.7} & 98.8 & 99.0 \\
			\bottomrule
		\end{tabular}	
		}
		\\[6pt]
		\resizebox{1\linewidth}{!}{%
		\begin{tabular}{l| c | c | c | c !{\vrule width 1.5pt} c |c | c }
			\multicolumn{8}{c}{\bfseries NME (\%) on CUB-200-2011 dataset $\downarrow$}  \\
			\toprule
			\bfseries Method     &  \multicolumn{7}{c}{\bfseries  Training set size}  \\
			&    1    &  10  & 20 &   50   &  5\% & 20\%  & 100\%  \\
			\toprule
			Xiao et al. \cite{xiao2018_simple-baseline} & 25.6 & 19.4 & 18.6 & 16.6 & 12.3 & 8.79 & 5.53\\
			Moskvyak et al. \cite{moskvyak2021_semi-supervised} & 64.1 & 55.2 & 51.8 & 39.5 & 12.3 & 7.42 & \textbf{3.95} \\
			AutoLink (few) \cite{he2022_autolink} & \textbf{20.7} & 18.1 & 16.8 & 14.4 & 10.6 & 8.72 & 5.45 \\
			\midrule
			ours & \textbf{20.7} & \textbf{9.94} & \textbf{9.17} & \textbf{8.97} & \textbf{6.42} & \textbf{5.19} & 4.58 \\
			\bottomrule
		\end{tabular}	
		}
		\\[6pt]
		\resizebox{1\linewidth}{!}{%
		\begin{tabular}{l| c | c | c | c !{\vrule width 1.5pt} c |c | c }
			\multicolumn{8}{c}{\bfseries NME (\%) on ATRW dataset $\downarrow$}  \\
			\toprule
			\bfseries Method     &  \multicolumn{7}{c}{\bfseries  Training set size}  \\
			&    1    &  10  & 20 &   50   &  5\% & 20\%  & 100\%  \\
			\toprule
			Xiao et al. \cite{xiao2018_simple-baseline} & 24.1 & 20.1 & 19.5 & 18.3 & 11.7 & 5.61 & 3.09 \\
			Moskvyak et al. \cite{moskvyak2021_semi-supervised} & 43.3 & 43.8 & 41.5 & 33.8 & 7.99 & 4.79 & 3.67 \\
			AutoLink (few) \cite{he2022_autolink} & 20.3 & 20.2 & 19.8 & 19.1 & 7.26 & 4.86 & 3.10 \\
			\midrule
			ours & \textbf{19.8} & \textbf{19.0} & \textbf{6.69} & \textbf{5.94} & \textbf{3.72} & \textbf{2.89} & \textbf{2.83} \\
			\bottomrule
		\end{tabular}	
		}
		\\[6pt]
		\resizebox{1\linewidth}{!}{%
		\begin{tabular}{l| c | c | c | c !{\vrule width 1.5pt} c |c | c }
			\multicolumn{8}{c}{\bfseries NME (\%) on CarFusion dataset $\downarrow$}  \\
			\toprule
			\bfseries Method     &  \multicolumn{7}{c}{\bfseries  Training set size}  \\
			&    1    &  10  & 20 &   50   &  5\% & 20\%  & 100\%  \\
			\toprule
			Xiao et al. \cite{xiao2018_simple-baseline} & 29.7 & 22.9 & 21.5 & 18.6 & 15.5 & 8.12 & 4.80 \\
			Moskvyak et al. \cite{moskvyak2021_semi-supervised} & 60.3 & 63.0 & 38.6 & 31.1 & 19.3 & 7.22 & 4.19 \\
			AutoLink (few) \cite{he2022_autolink} & \textbf{27.0} & 21.3 & 19.2 & 15.1 & 13.7 & 7.00 & 4.55 \\
			\midrule
			ours & 29.0 & \textbf{15.5} & \textbf{14.8} & \textbf{12.5} & \textbf{9.31} & \textbf{3.76} & \textbf{3.56} \\
			\bottomrule
		\end{tabular}
		}
		
	\end{center}
	\vspace{-0.5cm}
	\caption{\textbf{Additional Quantitative Comparison with Baselines on WFLW, SynthesEyes, CUB, ATRW, and CarFusion.}}
	\label{tab:quantitative_comparison_other}
\end{table}

\section{Negative Societal Impacts}
Our paper has no ethical concerns but might have some potential malicious misuses on downstream tasks, such as face tracking and animation. \looseness=-1

\section{Acknowledgments}
We thank Shih-Yang Su for his insightful comments on selecting core examples from a dataset, and anonymous reviewers for their valuable feedback.

\end{document}